\documentclass[conference]{IEEEtran}
\IEEEoverridecommandlockouts
\usepackage{cite}
\usepackage{amsmath,amssymb,amsfonts}
\usepackage{algorithmic}
\usepackage{graphicx}
\usepackage{textcomp}
\usepackage{xcolor}
\usepackage[T1]{fontenc}
\usepackage{cite}
\usepackage{url} 

\def\BibTeX{{\rm B\kern-.05em{\sc i\kern-.025em b}\kern-.08em
  T\kern-.1667em\lower.7ex\hbox{E}\kern-.125emX}}
\begin{document}

\title{Combining Forecasts using Meta-Learning:\\ A Comparative Study for Complex Seasonality\\
}

\author{\IEEEauthorblockN{Grzegorz Dudek}
\IEEEauthorblockA{\textit{Electrical Engineering Faculty} \\
\textit{Czestochowa University of Technology}\\
Częstochowa, Poland \\
grzegorz.dudek@pcz.pl}

}

\maketitle

\begin{abstract}
In this paper, we investigate meta-learning for combining forecasts generated by models of different types. While typical approaches for combining forecasts involve simple averaging, machine learning techniques enable more sophisticated methods of combining through meta-learning, leading to improved forecasting accuracy. We use linear regression, $k$-nearest neighbors, multilayer perceptron, random forest, and long short-term memory as meta-learners. We define global and local meta-learning variants for time series with complex seasonality and compare meta-learners on multiple forecasting problems, demonstrating their superior performance compared to simple averaging.
\end{abstract}

\begin{IEEEkeywords}
ensemble forecasting, machine learning, meta-learning, multiple seasonality, short-term load forecasting, stacking
\end{IEEEkeywords}

\section{Introduction}

Ensemble methods are widely recognized as a cornerstone of modern machine learning (ML) \cite{Ree18}, commonly used for regression and classification problems. In addition, ensembling has proven to be a highly effective approach for increasing the predictive power of forecasting models. The ensemble approach in forecasting, which involves combining the predictions of multiple models, can be justified for several reasons. First of all, it usually leads to increased accuracy. Ensemble models often outperform individual models, as they leverage the strengths of different models and minimize their weaknesses. By combining diverse models, the ensemble can produce more accurate predictions by capturing a broader range of patterns and insights from the data. Ensembling also allows for the incorporation of multiple drivers into the data generating process, mitigating uncertainties regarding model form and parameter specification \cite{Wan22}. This helps to reduce the risk of relying on a single model's limitations or biases, and enables a more comprehensive representation of the underlying data generating process.  

Ensembling is generally more robust in respect of outliers or extreme values in the data. While individual models may be overly influenced by outliers, an ensemble can mitigate this by averaging or combining predictions, resulting in a more stable and reliable forecast. Ensemble methods can reduce overfitting by combining the predictions of multiple models and reducing the risk of relying too heavily on one model's biases or noise. 

An ensemble approach can be flexible and adaptable, allowing for easy incorporation of new base models or modifications into existing models. This flexibility enables the ensemble to be updated or improved over time, leading to potentially better forecast performance. Finally, ensembling is often computationally efficient, as it can leverage parallel processing or other optimization techniques. This allows for faster prediction generation compared to more complex models. As a result, ensembling proves to be well-suited for real-time applications or scenarios demanding large-scale forecasting capabilities.

The effectiveness of ensembling in forecasting has been demonstrated in many forecasting competitions such as the M competitions. For instance, in the M4 competition, 12 out of the 17 most accurate models used some form of ensembling \cite{Ati20}. The winning submission in the M4 competition utilized three types of ensembling simultaneously, including combining the results of the stochastic training process, bagging, and combining multiple runs \cite{Smyl20}. 
The M5 competition confirmed the findings of previous M competitions by demonstrating that accuracy can be improved by combining forecasts obtained with different methods, even relatively simple ones \cite{Mak20}.

\subsection{Related Work}

There are numerous approaches for combining forecasts, with arithmetic average of forecasts based on equal weights being a popular and surprisingly robust method
that often outperforms more complicated weighting schemes \cite{Bla16,Gen13}. Other simple alternatives to the arithmetic average include combination strategies based on the median, mode, trimmed means, and winsorized means \cite{Jos08,Lic20}. These methods may perform even better than the arithmetic average because they are more robust, meaning they are less sensitive to extreme forecasts. All these simple methods are easy to implement, not computationally burdensome and can be quickly deployed in practical forecasting scenarios. The theoretical properties of forecast combination, as investigated in \cite{Cha18}, shed light on why a simple average of forecasts often outperforms forecasts from single models.

To differentiate the weights assigned to individual models, linear regression can be used. In this case, the combination weights can be estimated using ordinary least squares, where the vector of past observations is used as the response variable and the matrix of past individual forecasts serves as the predictor variables. The weights can reflect historical performances of the base models, such as in \cite{Paw20}, where the weights inversely proportional to the prediction error were considered. In situations where multicollinearity of individual forecasts is present, principal components regression can be employed \cite{Pon11}. Additionally, forecast combinations using changing weights have been developed to address different types of instabilities in constituent forecasts \cite{Ros21}.
Furthermore, weights can also be derived from information criteria, such as AIC \cite{Kol11}. In \cite{Kan22}, the weights were estimated based on the diversity of individual learners.
In \cite{Fu22}, a more sophisticated framework for determining the weights was proposed using a reinforcement learning based model while \cite{Mon20} introduced a FFORMA approach, which obtains the weights using time series characteristic features.

If the constituent forecasts are derived from nonlinear models, or if the true relationship between combination members and the best forecast is characterized by nonlinear systems, ML models can be utilized to nonlinearly combine the base forecasts using a stacking procedure \cite{Bab16,Gas21}. The stacking approach can improve forecast accuracy by learning the optimal combination of constituent forecasts in a data-driven manner. 

There are many examples in the literature showing the advantages of  the stacking generalization. From the perspective of this work, we focus on stacking for time series with complex seasonality. Ribeiro et al. \cite{Rib19} proposed a method to handle time series with multiple seasonal cycles by using wavelet neural networks (WNNs) as both base forecasters and a meta-learner. This approach improves the accuracy of the forecasts while maintaining the ability to capture different seasonal patterns in the data. 
In \cite{Div18}, the authors used regression trees, random forest, and neural networks as base models for short-term electrical consumption forecasting with triple seasonal patterns. An algorithm based on gradient boosting was used as a meta-learner. For a similar forecasting problem, \cite{Suj22} combined four base models, i.e. random forests, long short-term memory, deep neural networks, and evolutionary trees, with gradient boosting and extreme gradient boosting as meta-learners, demonstrating a significant reduction in forecast error.
In \cite{Gas21}, the authors compared different strategies of combining forecasts, including simple average, linear combination with weights based on performance, FFORMA, and stacking. Experimental results on 16,000 time series from various sources showed that stacking outperforms its competitors, highlighting its effectiveness for time series with different characteristics.

In stacking, the base models are typically of different types and fit on the same dataset. The diversity of the ensemble results from the different properties of the base models. 
On the other hand, bagging, another popular ensemble learning approach, achieves diversity by training models, usually of the same type, on different subsets of the training set, with each model's results being equally considered. 
Random forest is a typical example of the bagging approach, combining many decision trees as base learners, with each tree trained on a bootstrap sample of the training set. An example of using  bagging based on random forest for time series with complex seasonality is \cite{Dud22}. In this study, to deal with multiple seasonal cycles, several methods of time series preprocessing and training modes are proposed.  
In \cite{Khw15}, the bagging ensemble employs multilayer perceptrons while in \cite{Don17}, it employs convolutional neural networks. Both papers report improvement in short-term load forecasting accuracy.

Boosting is another powerful method for combining base learners. In this approach, a sequence of base models of the same type is generated to correct the predictions of prior models. These models are iteratively trained to capture the patterns in the residuals of the previous iteration, resulting in a sequence of increasingly accurate predictions. Gradient-boosted trees, such as the extreme gradient boosting model, are popular representatives of boosting. These were used for short-term load forecasting in \cite{Wan21} and \cite{Zha20}, and the results demonstrate the efficiency of the proposed boosting strategies, showing an improvement in accuracy over baseline models. In \cite{Dud22a}, new boosting strategies based on corrected targets and opposed responses were proposed for the same forecasting problem with the aim of unifying the forecasting tasks for all learners, leading to simplified ensemble learning and increased forecasting accuracy. The stacking, bagging and boosting approaches were compared in \cite{Rib20}.  

\subsection{Motivation and Contribution}

In this study, we propose a stacking approach that utilizes different ML models as meta-learners to combine forecasts generated by individual forecasting models of different types. We aim to test various ML models that employ different strategies to solve a regression problem involving the modeling the target variable (forecast) based on forecasts generated by the base models. Namely, the linear regression model combines the base forecasts in a linear manner, with weights determined on historical data. The $k$-nearest neighbor model represents a non-parametric regression function that averages the target forecasts of the most similar training patterns to the query pattern. A multilayer perceptron constructs a model that combines nonlinear projections of the base forecasts. A random forest model combines the base forecasts using regression trees and bagging strategy. Lastly, a long short-term memory (LSTM) model creates a nonlinear regression function with memory, allowing the model to capture temporal relationships in the data.

We consider both global and local learning strategies in our stacking approach. The global approach utilizes all historical data for training the meta-models, while the local approach trains the meta-models on the most similar patterns to the query pattern. In the case of LSTM, the local approach limits the training patterns to the recent ones or selects them based on seasonality in the time series. 

The contribution of this study can be summarized as follows:
\begin{enumerate}
\item 
We explore the use of ML models of various types as meta-learners for combining base forecasts generated by a set of forecasting models. This allows us to investigate and assess the effectiveness of different ML models in capturing patterns and insights from the base forecasts and producing more accurate combined forecasts.
\item 
We investigate global and local strategies of meta-learning for time series with complex seasonality. By considering both global and local approaches, we aim to capture both the overall patterns and local nuances in the data, which can be crucial for accurate forecasting in time series with complex seasonality.
\item 
We conduct experiments on 35 time series with triple seasonality, using 16 base models, to validate the efficacy of the proposed approach. The experimental results demonstrate the high performance of the proposed meta-learners in combining forecasts more accurately than simple averaging.
\end{enumerate}

The rest of the work is organized as follows. Section II describes the proposed meta-learners. The global and local meta-learning variants are presented in Section III. Section IV gives some application examples and discusses the results. Finally, Section V concludes the work.

\section{Meta-learners}

The problem of forecast combinations involves the task of finding a regression function, denoted as $f$, that aggregates forecasts for time $t$ generated by $n$ forecasting models. The function can utilize all available information up to time $t-h$, where $h$ represents the forecast horizon. However, in this study, we restrict this information to the base forecasts represented by vector $\hat{\textbf{y}}_t=[\hat{y}_{1,t},..., \hat{y}_{n,t}]$. The combined forecast is given by $\tilde{y}_t=f(\hat{\textbf{y}}_t; \boldsymbol{\theta}_t)$, where $\boldsymbol{\theta}_t$ represents meta-model parameters.

The class of regression functions $f$ encompasses a range of mappings, including both linear and nonlinear ones. The meta-model parameters can be either static or time-varying throughout the forecasting horizon. To optimize the performance of the ensemble, we adopt an approach where the parameters are learned individually for each forecasting task, using a specific training set tailored for that task: $\Phi=\{(\hat{\textbf{y}}_\tau,y_\tau)\}_{\tau \in \Xi}$, where $y_\tau$ represents the target value and $\Xi$ represents a set of selected time indexes from interval $T=1, ..., t-h$. The process of selecting this set is elaborated in Section III.

In this study, we propose five meta-models whose main properties, advantages and disadvantages are characterized below.

\subsection{Linear Regression (LR)}

LR combines the base forecasts as follows:  

\begin{equation}
 f(\hat{\textbf{y}}) = \sum_{i=1}^n{a_i\hat{y}_i} + a_0
\end{equation}
where $a_0, ..., a_n$ are coefficients.

LR is simple, easy to interpret and computationally efficient but it has limited flexibility as it assumes a linear relationship between the base forecasts and the target forecast. Moreover, it is sensitive to outliers and influential observations, and cannot handle multicollinearity between the independent variables.

LR assumes that the model is time independent, and the coefficients estimated on historical data will be appropriate for future data. So for future data, far from the moment when the parameters of the model are estimated, the model may become outdated. To prevent this, in this study, we estimate the parameters individually for each forecasting task (test point). This also applies to the other models described below.  

\subsection{$k$-Nearest Neighbours (kNN)}

The kNN method is a popular technique for regression analysis that falls under the category of instance-based learning. 
In kNN regression, the goal is to estimate the value of a continuous target variable for a new observation based on the $k$ nearest neighbors to this observation in the training data. 
The prediction is based on the average or weighted average of the target values of these $k$ nearest neighbors, which makes the model suitable for capturing local patterns and trends in the data.

kNN is a non-parametric method, meaning it does not make any assumptions about the underlying distribution of the data. It does not fit a specific functional form or estimate model parameters, which makes it flexible and versatile for a wide range of data distributions. The kNN regression function in our implementation is as follows:

\begin{equation}
 f(\hat{\textbf{y}}) = \frac{\sum_{\tau \in \Xi}{w(\hat{\textbf{y}},\hat{\textbf{y}}_{\tau})y_{\tau}}}{\sum_{\tau \in \Xi}{w(\hat{\textbf{y}},\hat{\textbf{y}}_{\tau})}}
\end{equation}
where
\begin{equation}
 w(\hat{\textbf{y}},\hat{\textbf{y}}_{\tau}) = \exp{\left(-\frac{\|\hat{\textbf{y}}-\hat{\textbf{y}}_{\tau}\|^2}{\sigma^2}\right)}
\end{equation}
is the Gaussian-type weighting function, $\|.\|$ denotes Euclidean norm and $\sigma$ is a bandwidth parameter.

kNN is a simple and intuitive method to implement. The algorithm involves only two main steps: finding the $k$ nearest neighbors of a new observation and calculating the prediction based on the target variable values of these neighbors. 
Hyperparameters $k$ and $\sigma$ both decide about the bias-variance trade-off. The higher their values, the smoother the regression function, which means lower variance but increased bias.

\subsection{Multilayer Perceptron (MLP)}

MLP is a popular type of neural network that is commonly used for regression problems. This is because it has many beneficial properties, such as its ability to approximate any function, learn from data, model nonlinear relationships, process data in parallel, tolerate noise, and tolerate faults.

The proposed MLP architecture in this study consists of one hidden layer with hyperbolic tangent nonlinearities, $n$ inputs, and a single output node. The regression function of MLP can be expressed as follows:

\begin{equation}
 f(\hat{\textbf{y}}) = \sum_{j=1}^m{v_j\phi_j(\hat{\textbf{y}})} + v_0
 \label{eq2}
\end{equation}
where 
\begin{equation}
\phi_j(\hat{\textbf{y}})=\frac{2}{1+\exp{\left(-\left(\sum_{i=1}^n{w_{i,j}\hat{y}_i+w_{0,j}}\right)\right)}}-1
 \label{eq3}
\end{equation}
$m$ denotes the number of hidden nodes, $w$ and $v$ are the weights of the hidden and output layer, respectively.

As can be seen from \eqref{eq2} and \eqref{eq3}, the base forecasts are first nonlinearly projected into $m$-dimensional space, and then combined linearly by the output node. The MLP's approximation abilities are linked to the number of hidden nodes, $m$. To avoid overfitting, we train MLP using the Levenberg-Marquardt algorithm, along with Bayesian regularization, which minimizes the sum of squared errors and weights.

\subsection{Random Forest (RF)}

RF is a type of ensemble learning that employs decision trees as base models, as proposed by Breiman in \cite{Bre01}. For regression problems, we utilize regression RF, which constructs and aggregates multiple regression trees. The method uses bagging in combination with a random subspace technique to create a set of base models that are noisy yet almost unbiased, thereby reducing variance. To enhance diversity among base models, a random subspace method is employed to limit trees to operate on distinct random subspaces of the predictor space. Additionally, bagging boosts diversity by constructing each tree in the forest from a bootstrap sample of the original dataset.

The RF model is expressed as follows:

\begin{equation}
 f(\hat{\textbf{y}}) = \frac{1}{p}\sum_{j=1}^p{T_j(\hat{\textbf{y}}})
 = \frac{1}{p}\sum_{j=1}^p\sum_{l \in L_j} label(l)I(\hat{\textbf{y}} \in l)
 \label{eqdt}
\end{equation}
where $p$ is the number of trees in a forest, $T_j(\hat{\textbf{y}})$ is a response of the $j$-th tree on the query pattern $\hat{\textbf{y}}$,
$L_j$ is a set of leaves of the $j$-th tree, $label(l)$ is a function, which assigns a label to leaf $l$ based on the subset of samples that reached that leaf (typically the label is the average of the responses of these samples), and $I(\hat{\textbf{y}} \in l)$ returns 1 when sample $\hat{\textbf{y}}$ reaches leaf $l$, and 0 otherwise.

The function modelled by the tree, $T_j(\hat{\textbf{y}})$, is a step function. Aggregating multiple trees in the forest smooths the regression function by reducing the step size.

A decision tree is defined by various parameters, such as split predictors and cutpoints at each node, and terminal-node (leaf) values calculated by function $label(l)$. These parameters are determined during the learning process using a split criterion, which is typically a mean square error for regression tasks. The primary hyperparameters of RF include the number of trees in the forest, $p$, the minimum number of observations in a leaf (or equivalent measure), $q$, and the number of predictors randomly selected for each split, $r$. All hyperparameters affect the trade-off between bias and variance of the model.

\subsection{Long Short-Term Memory (LSTM)}

LSTM is a modern recurrent NN that incorporates a gating mechanism \cite{Hoc97}. This NN architecture was specifically designed to handle sequential data and is capable of learning short and long-term relationships in time series \cite{Hew21}. LSTM is composed of recurrent cells that can maintain their internal states over time, i.e. cell state \textbf{c} and hidden state \textbf{h}. These cells are regulated by a nonlinear gating mechanism that controls the flow of information within the cell, allowing it to adapt to the dynamics of the modeled process.

In our implementation, the LSTM network consists of two layers: the LSTM layer and the linear layer. The LSTM layer is responsible for approximating temporal dependencies in sequential data and generating state vectors, while the linear layer converts hidden state vector \textbf{h} into the output value. The function modeled by the LSTM network can be written as:

\begin{equation}
 f(\hat{\textbf{y}}_t) = \textbf{v}^T\textbf{h}_t(\hat{\textbf{y}}_t) + v_0
 \label{eqls}
\end{equation}
where $\textbf{h}_t(\hat{\textbf{y}}_t)=\text{LSTM}(\hat{\textbf{y}}_t, \textbf{c}_{t-1}, \textbf{h}_{t-1}; \textbf{w}) \in \mathbb{R}^{m}$, $\textbf{w}$ and $\textbf{v}$ are the weights of the LSTM and linear layers, respectively.

The number of nodes in each gate, $m$, is the most critical hyperparameter. It determines the amount of information stored in the states. For more intricate temporal relationships, a higher number of nodes is necessary.

In contrast to the other ensemble models examined in this study, 
to calculate output $\tilde{y}_t$, LSTM uses not only the information included in the base forecasts for time $t$, $\hat{\textbf{y}}_t$, but also the information in the base forecasts for previous time steps, $t-1, t-2, ...$. 
This is achieved through states $\textbf{c}_{t-1}$ and $\textbf{h}_{t-1}$, which accumulate and store  information from the past steps. The ability to incorporate past information allows LSTM to capture complex patterns and dependencies in the sequential data.

\section{Meta-learning Variants}

The base forecasting models generate forecasts for successive time points $T=1, ..., t$. To obtain an ensemble forecast for time $t$, we can train the meta-model using all available historical data from period $\Xi={\{1, ..., t-h\}}$, which is referred to as the global approach. Using this method, the model can utilize all available past information to generate a forecast for the current time point $t$.

In the local mode, we aim to learn combining function $f$ locally around query pattern $\hat{\textbf{y}}_t$. To achieve this, we select the $k$ most similar input vectors to $\hat{\textbf{y}}_t$ and include them in the training set.  We apply this approach to non-recurrent meta-models, using the Euclidean metric to determine the nearest neighbors.
However, for LSTM, which captures dynamics from sequential data, this approach can disrupt the temporal structure of the data. To address this issue, we define local learning differently for LSTM. Specifically, we restrict the training sequence to the last $c$ points, i.e. $\Xi={\{t-h-c, ..., t-h\}}$, allowing the LSTM to model the relationship for query pattern $\hat{\textbf{y}}_t$ based on the most recent sequence of length $c$. We refer to this approach as v1.

For seasonal time series, it may be beneficial to select training points for LSTM that are lagged to the forecasted point by the length of the seasonal period $s_1$. In such a case, the training set is composed of points $\Xi=\{t-cs_1, t-(c-1)s_1, ..., t-s_1\}$, where $c$ is the  size of the training set.
Note that this training set preserves the time structure of the data, but with a modified version that disregards the seasonal pattern. Specifically, it includes only those points that are in the same phase of the seasonal cycle as the forecasted point. This approach is referred to as v2.

\begin{figure}[h]
	\centering
	\includegraphics[width=0.5\textwidth]{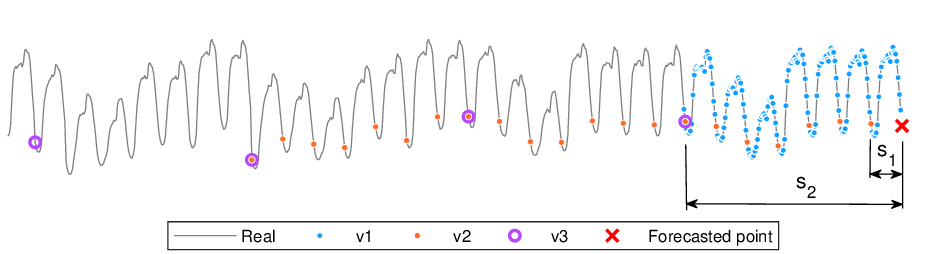}
	\caption{Training points for LSTM in variants v1, v2 and v3.} 
	\label{fig0}
\end{figure}

In the case of time series with double seasonality, where the periods are denoted as $s_1$ and $s_2$ (with $s_2$ being a multiple of $s_1$), a modified training set for LSTM contains points from the same phase of both seasonal patterns as the forecasted point. Specifically, the training set is composed of points $\Xi={t-cs_2, t-(c-1)s_2, ..., t-s_2}$. We refer to this approach as v3. Fig. \ref{fig0} provides a visualization of the training points for each variant of LSTM training.

It is important to note that approaches v2 and v3 remove points from the time series that are not in the same phase as the forecasted point. This simplifies the relationship between the new training points and the forecasted point, making it easier to model. However, this simplification comes at the cost of potentially losing some of the information related to the seasonal patterns that occur outside of the selected phase. Therefore, it is important to carefully consider which approach to use depending on the specific characteristics of the data.

\section{Experimental Study}

In this section, we evaluate the performance of our proposed stacking approaches for combining forecasts. We consider a short-term load forecasting problem for 35 European countries. The time series exhibit triple seasonality, i.e., daily, weekly, and yearly. The base models comprise 16 forecasting models of different types, which are detailed in Section IV-B. 

\subsection{Dataset, Training and Evaluation Setup}

We collected real-world data from the ENTSO-E repository (\url{www.entsoe.eu/data/power-stats}) to use in our study. The dataset consists of hourly electricity loads recorded from 2006 to 2018, covering 35 European countries. The dataset offers a diverse range of time series, each exhibiting unique characteristics such as levels and trends, variance over time, intensity and regularity of seasonal fluctuations spanning different periods (annual, weekly, and daily), and varying degrees of random fluctuations.

The forecasting base models were optimized and trained on data from 2006 to 2017 and applied to generate hourly forecasts for the year 2018, on a daily basis (see \cite{Smy23} for details). To evaluate the performance of the meta-models, 100 hours were selected for each country from the second half of 2018 (evenly spaced across the period), and the forecasts for each of these hours were combined by meta-models. The meta-models were trained separately for each selected hour, using data from January 1, 2018 up to the hour preceding the forecasted hour ($h=1$) for optimization and training. For LSTM variant v2, we assumed a daily seasonality period of $s_1=24$ hours, while for variant v3, we assumed a weekly period of $s_2=7\cdot24=168$ hours (see Fig. \ref{fig0}).

The meta-models were trained in the global mode (on all $N_t$ available historical datapoints) and local modes. The local modes for RL, kNN, MLP and RF refer to learning on the $k$ nearest training patterns to the query pattern, where $k \in K=\{20, 40, ..., 200, 250, 300\}$. The local mode for LSTM refers to learning on the $c$ last training patterns, where $c \in C=\{24, 48, 72, 168, 504\}$, i.e. on a training period ranging from one day to three weeks.     

We evaluate the meta-models for their different hyperparameter values as follows:
\begin{itemize}
\item LR has no hyperparameters.
\item Hyperparameter $k$ in kNN was searched over the set $K+\{N_t\}$.  
We made the bandwidth dependent on the data by calculating it as a function of the median of distances $d$ between the query pattern and the training patterns: $\sigma=b \cdot \text{median}(d)$. Hyperparameter $b$ was searched over the set $\{0.03, 0.05, 0.07\}$  selected by experimetation.
\item Number of the hidden neurons in MLP was searched over the set $\{1, 3, 5\}$ selected in preliminary tests and the number of training epochs was set at 100.
\item For RF, we select default values for hyperparameters after preliminary simulations confirming their good performance: number of predictors to select at random for each decision split $r=n/3$ as recommend by the RF inventors, minimum number of observations per tree leaf $q=1$ (this produces overtrained trees, but combining them in the forest reduces overtraining), number of trees in the forest $p=100$.
\item Some of the LSTM hyperparameters were set to default values (we used Matlab implementation of LSTM), while others were determined through experimentation. They include the number of nodes, which was set at $m=128$, and the number of epochs, which was set at 200. 
\end{itemize}

The proposed meta-models were implemented using Matlab 2022b. The experiments were conducted on a Microsoft Windows 10 Pro operating system, with an Intel(R) Core(TM) i7-6950x
CPU @3.0 GHz processor and 48 GB RAM.

The performance of the models was evaluated using the following metrics:
MAPE -- mean absolute percentage error, MdAPE -- median of absolute percentage error, MSE -- mean squared error, MPE -- mean absolute percentage error, and StdPE -- standard deviation of percentage error. These metrics provide a comprehensive assessment of the accuracy and precision of the forecasting models.

\subsection{Base Forecasting Models}

We employed a diverse set of forecasting models as the base models for our study. These include statistical models, classical ML models, as well as recurrent, deep, and hybrid NN architectures sourced from \cite{Smy23}. This comprehensive selection covers a wide range of modeling techniques with different  mechanisms for capturing temporal patterns in data. By incorporating these various models, we aimed to ensure sufficient diversity in the base learners to improve collective forecasting performance.  

\begin{itemize}
\item ARIMA -- autoregressive integrated moving average model,
\item ES -- exponential smoothing model,
\item Prophet -- modular additive regression model with nonlinear trend and seasonal components,
\item N-WE -- Nadaraya–Watson estimator,
\item GRNN -- general regression NN,
\item MLP -- perceptron with a single hidden layer and sigmoid nonlinearities,
\item SVM -- linear epsilon insensitive support vector machine ($\epsilon$-SVM),
\item LSTM -- long short-term memory,
\item ANFIS -- adaptive neuro-fuzzy inference system,
\item MTGNN -- graph NN for multivariate time series forecasting,
\item DeepAR -- autoregressive recurrent NN model for probabilistic forecasting,
\item WaveNet -- autoregressive deep NN model combining causal filters with dilated convolutions,
\item N-BEATS -- deep NN with hierarchical doubly residual topology,
\item LGBM -- Light Gradient-Boosting Machine,
\item XGB -- eXtreme Gradient Boosting algorithm,
\item cES-adRNN -- contextually enhanced hybrid and hierarchical model combining exponential smoothing and dilated recurrent NN with attention mechanism.
\end{itemize}

See \cite{Smy23} for more information on base models.

\subsection{Results and Discussion}

Table \ref{tabEr} shows the results of the base models averaged over 100 selected hours from 2018 and 35 countries. 
As can be seen from this table, the models vary in MAPE from 1.70 to 3.83, and in MSE from 224,265 to 1,641,288. The most accurate model in terms of MAPE, MdAPE and MSE is cES-adRNN, while the least accurate is Prophet. 

\begin{table}[htbp]
\setlength{\tabcolsep}{6pt}
\caption{Forecasting quality metrics for the base models.}
\begin{center}
\begin{tabular}{|l|r|r|r|r|r|}
\hline
 Model & MAPE & MdAPE & MSE & MPE & StdPE 
\\
\hline
ARIMA & 2.86 & 1.82 & 777012 & 0.0556 & 4.60 \\
ES  & 2.83 & 1.79 & 710773 & 0.1639 & 4.64 \\
Prophet  & 3.83 & 2.53 & 1641288 & -0.5195 & 6.24 \\
N-WE & 2.12 & 1.34 & 357253 & 0.0048 & 3.47 \\
GRNN & 2.10 & 1.36 & 372446 & 0.0098 & 3.42 \\
MLP  & 2.55 & 1.66 & 488826 & 0.2390 & 3.93 \\
SVM  & 2.16 & 1.33 & 356393 & 0.0293 & 3.55 \\
LSTM & 2.37 & 1.54 & 477008 & 0.0385 & 3.68 \\
ANFIS & 3.08 & 1.65 & 801710 & -0.0575 & 5.59 \\
MTGNN & 2.54 & 1.71 & 434405 & 0.0952 & 3.87 \\
DeepAR  & 2.93 & 2.00 & 891663 & -0.3321 & 4.62 \\
WaveNet  & 2.47 & 1.69 & 523273 & -0.8804 & 3.77 \\
N-BEATS  & 2.14 & 1.34 & 430732 & -0.0060 & 3.57 \\
LGBM & 2.43 & 1.70 & 409062 & 0.0528 & 3.55 \\
XGB  & 2.32 & 1.61 & 376376 & 0.0529 & 3.37 \\
cES-adRNN & 1.70 & 1.10 & 224265 & -0.1860 & 2.57 \\
\hline
\end{tabular}
\label{tabEr}
\end{center}
\end{table}

\begin{figure}[h]
\centering
\includegraphics[width=0.5\textwidth]{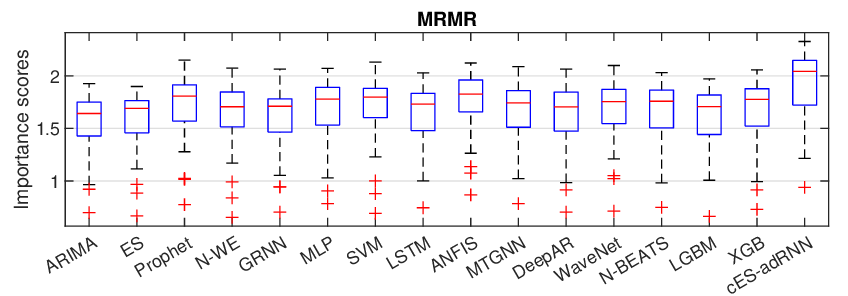}
\includegraphics[width=0.5\textwidth]{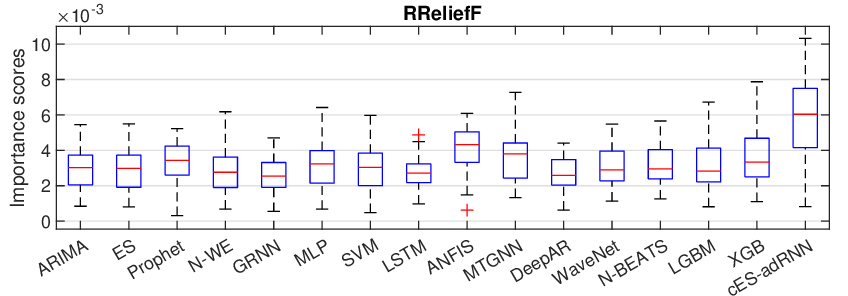}
\caption{Importance scores of the forecasting models determined using MRMR and RReliefF methods.} 
	\label{fig2}
\end{figure}

Treating the forecasts generated by the base models as input variables and the target value as the response variable, the importance scores of the inputs are calculated using two methods. The minimum redundancy maximum relevance method (MRMR) identifies a set of inputs that are both dissimilar to each other and effective in representing the response variable \cite{Din05}. On the other hand, the RReliefF method uses a nearest-neighbor approach to determine the relevance of inputs based on their contribution to the correct response of the nearest neighbors \cite{Rob97}.
The importance scores obtained using both methods are illustrated in Fig. \ref{fig2}. Notably, the most accurate model, cES-adRNN, is identified as the most important input. Surprisingly, ANFIS, one of the least accurate models, is ranked as the second most important input by both algorithms. Based on these results, we can select specific base models for the ensemble. However, in this study, we opted to include all 16 models in the ensemble, without explicitly selecting a subset of base models.

\begin{figure}[h]
	\centering
	\includegraphics[width=0.187\textwidth]{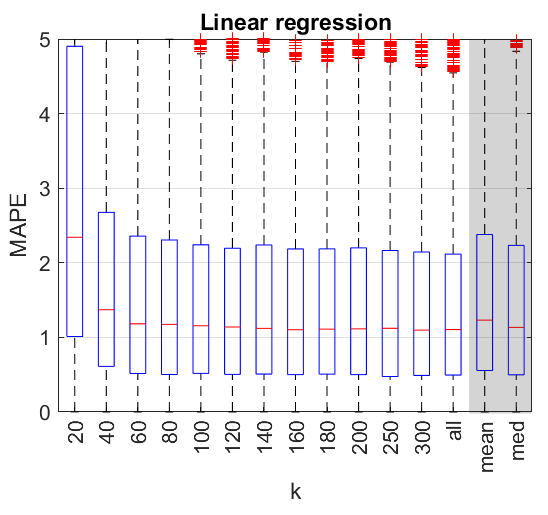}
  	\includegraphics[width=0.187\textwidth]{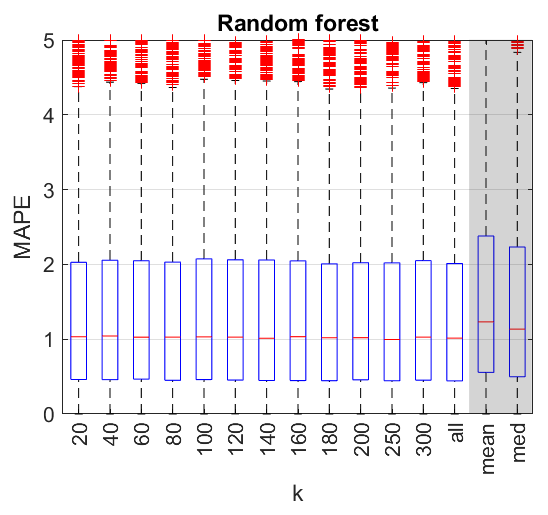}
    \includegraphics[width=0.49\textwidth]{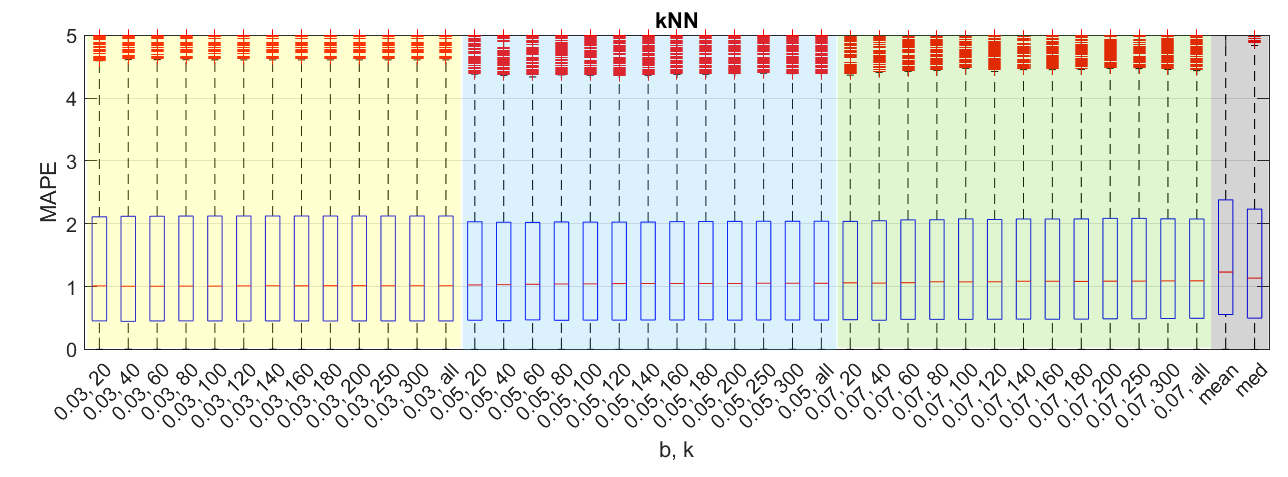}
	\includegraphics[width=0.49\textwidth]{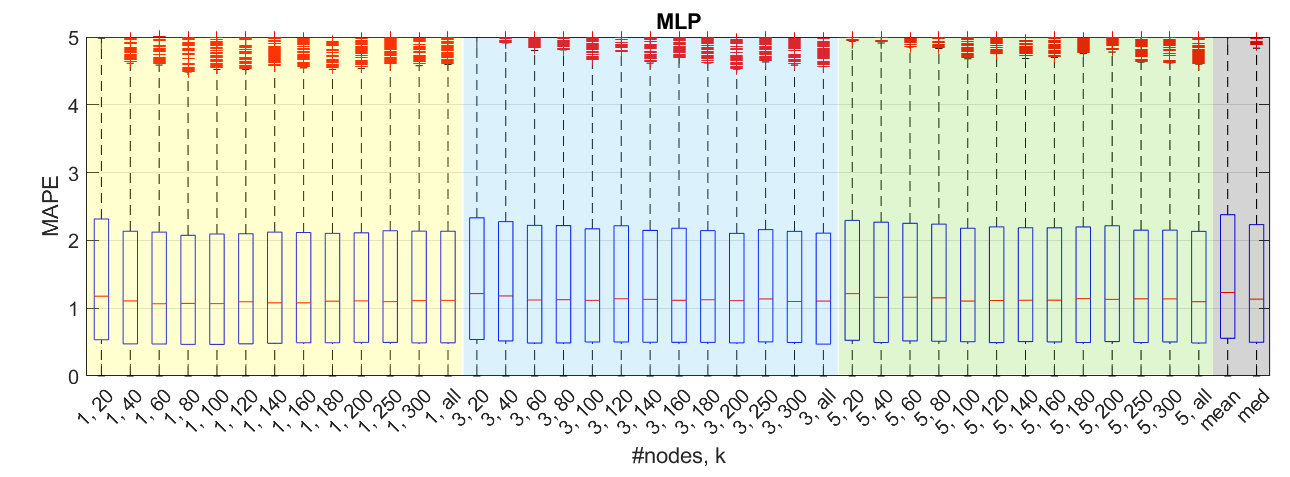}
	\includegraphics[width=0.3090\textwidth]{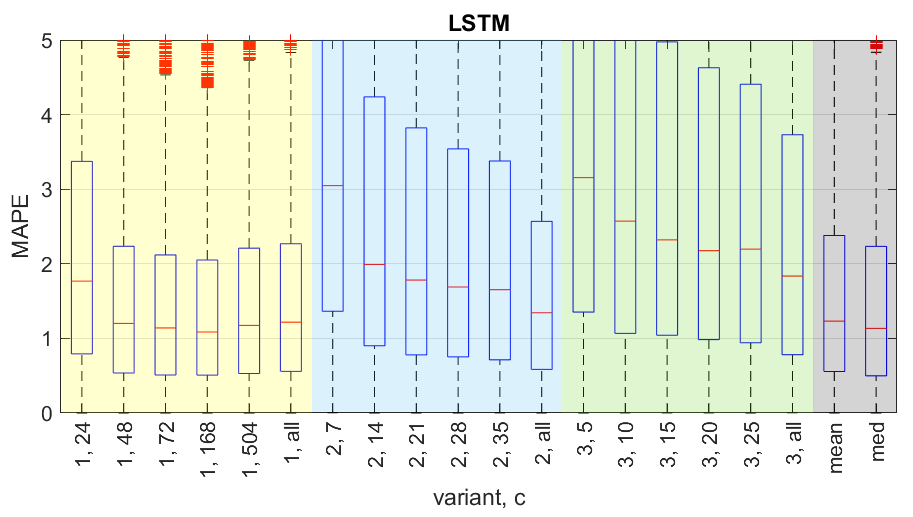}
	\caption{MAPE boxplots for the various ensemble variants.} 
	\label{figP}
\end{figure}

Fig. \ref{figP} shows the forecasting errors (MAPE) for different meta-learning variants and hiperparameters of the meta-models. 
Mean and Median, which represent the mean and median of the 16 forecasts produced by the base models, are also included for comparison. The following conclusions can be drawn from this figure:

\begin{itemize}
\item LR is observed to be the most sensitive to the size of the training set among the non-recurrent meta-models. It demonstrates the highest accuracy when trained on the entire training dataset (global mode).
\item kNN is found to be almost insensitive to the number of nearest neighbors $k$, meaning that varying the value of $k$ does not significantly impact its performance. Additionally, kNN exhibits little sensitivity to bandwidth parameter $b$ within the considered interval.
\item MLP tends to produce slightly less accurate predictions when trained on small datasets. However, as the training set size increases, its accuracy improves and stabilizes. Interestingly, the best results were achieved when using only one hidden neuron, indicating that the underlying relationship being modeled exhibits a relatively small degree of non-linearity.
\item RF exhibits low sensitivity to the size of the training set. It achieves its lowest error when trained in the global learning mode. However, even when the size of the training set is significantly reduced, the model's performance does not deteriorate significantly. This highlights the robustness of RF in handling different training set sizes and its ability to provide reliable predictions even with limited data.
\item LSTM in variants v2 and v3 demonstrates high sensitivity to the length of the training sequence. The model achieves the lowest errors when trained using the entire available data. Extending the training sequence has the potential to further reduce errors for these variants. However, LSTM v1 exhibits different behavior, with training sequences of length 168 hours (equivalent to one week) resulting in the lowest errors. Contrary to expectations, variants v2 and v3, which were designed to handle seasonal time series forecasting better, do not perform as well and show higher errors compared to the v1 variant.
\end{itemize}

Table \ref{tabEr2} provides a summary of the results, displaying the quality metrics for the best variants of the meta-models. From the table, it can be observed that RF performs the best in terms of MAPE and MdAPE, while LSTM achieves the lowest MSE. However, kNN and MLP are also competitive in terms of accuracy. LR performs slightly worse in accuracy measures but excels in forecast bias, as indicated by the MPE. It is worth noting that all the proposed meta-learners outperform the Median and Mean methods in terms of both forecast accuracy and dispersion, as measured by StdPE. Comparing the results from Table \ref{tabEr2} with those from Table \ref{tabEr}, it is evident that the proposed meta-models consistently produce more accurate predictions compared to the base models. 

\begin{table}[htbp]
\setlength{\tabcolsep}{3pt}
\caption{Forecasting quality metrics for different meta-models.}
\begin{center}
\begin{tabular}{|l|l|c|c|c|c|c|}
\hline
  & Variant  & MAPE & MdAPE & MSE & MPE & StdPE \\
\hline
Mean  & - & 1.91 & 1.23 & 316943 & --0.0775 & 3.11 \\
Median & - & 1.82 & 1.13 & 287284 & --0.0682 & 3.05 \\
LR & global & 1.63 & 1.11 & 213428 & \textbf{0.0131} & 2.38 \\
kNN & $k=40$, $b=0.05$,  & 1.54 & 1.03 & 178699 & --0.0915 & 2.33 \\
MLP  & $k=120$, $\text{\#nodes}=1$ & 1.59 & 1.09 & 180839 & --0.0786 & 2.31 \\
RF & global & \textbf{1.52} & \textbf{1.01} & 173821 & --0.0837 & \textbf{2.26} \\
LSTM  & v1, $c=168$  & 1.55 & 1.09 & \textbf{139667} & 0.0247 & \textbf{2.26} \\
\hline
\end{tabular}
\label{tabEr2}
\end{center}
\end{table}

\begin{figure}[h]
\centering
\includegraphics[width=0.49\textwidth]{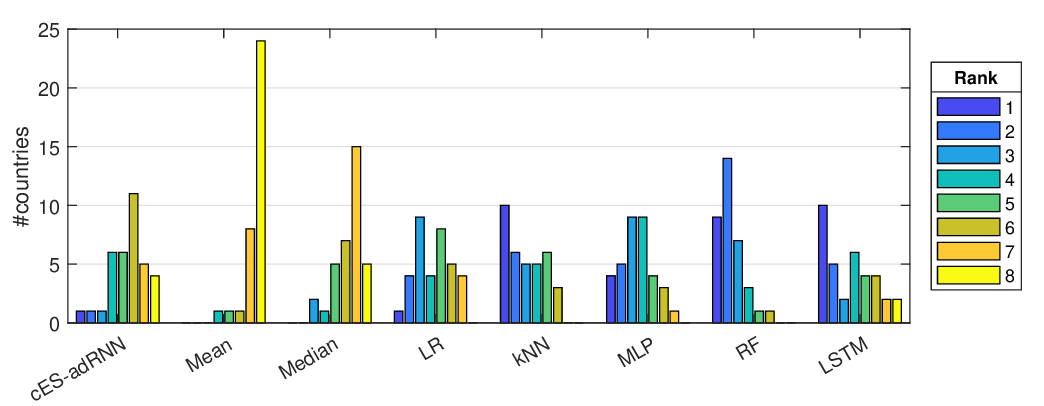}
\caption{MAPE ranking.} 
\label{fig3}
\end{figure}

Table \ref{tabP} provides a comprehensive breakdown of the results for each country, comparing the MAPE scores of the proposed meta-models with the most accurate base model, cES-adRNN. It is evident that cES-adRNN achieved the lowest MAPE for only one country, DK. LR had a similar performance, while the Mean and Median methods did not achieve the best score for any country. The model ranking based on MAPE is visually represented in Fig. \ref{fig3}. It shows that kNN, LSTM, and RF consistently occupied the top positions, with kNN and LSTM leading in the first position most frequently (10 times), followed closely by RF, which achieved the first position 9 times and the second position 14 times.
On the other hand, the Mean method was consistently ranked last (24 times) in terms of MAPE.

\begin{table}[htbp]
\setlength{\tabcolsep}{3pt}
\caption{MAPE for each country.}
\begin{center}
\begin{tabular}{@{}|c|c|c|c|c|c|c|c|c|@{}}
\hline
Country & cES-adRNN   & Mean & Median & LR & kNN  & MLP  & RF & LSTM \\ 
\hline
AL  & 2.12 & 2.15 & 2.02 & \textbf{1.95} & 2.09 & 2.12 & 2.08 & 2.02 \\
AT  & 1.50 & 1.90 & 1.77 & 1.50 & 1.37 & 1.41 & \textbf{1.33} & 1.44 \\
BA  & 1.37 & 1.56 & 1.56 & 1.35 & 1.49 & 1.34 & 1.41 & \textbf{1.30} \\
BE  & 2.58 & 2.88 & 2.91 & 2.61 & \textbf{2.10} & 2.75 & 2.53 & 2.25 \\
BG  & 1.52 & 1.65 & 1.58 & 1.49 & \textbf{1.34} & 1.50 & 1.39 & 1.72 \\
CH  & 2.46 & 2.67 & 2.57 & 2.47 & 2.66 & 2.68 & \textbf{2.37} & 2.98 \\
CZ  & 1.09 & 1.45 & 1.32 & 0.96 & 1.08 & 1.05 & \textbf{0.90} & 1.09 \\
DE  & 1.10 & 1.39 & 1.23 & 1.12 & 0.98 & \textbf{0.92} & 1.01 & 1.16 \\
DK  & \textbf{1.50} & 1.89 & 1.74 & 1.56 & 1.80 & 1.72 & 1.55 & 1.89 \\
EE  & 1.68 & 1.84 & 1.67 & 1.46 & \textbf{1.23} & 1.47 & 1.36 & 1.31 \\
ES  & 1.08 & 1.40 & 1.31 & 0.94 & \textbf{0.81} & 0.85 & 0.86 & 0.98 \\
FI  & 1.15 & 1.31 & 1.29 & 1.15 & \textbf{0.99} & 1.16 & 1.06 & 1.19 \\
FR  & 1.43 & 1.90 & 1.71 & 1.31 & 1.38 & 1.32 & 1.36 & \textbf{1.29} \\
GB  & 3.04 & 3.24 & 3.27 & 3.19 & 2.90 & 3.08 & 2.84 & \textbf{2.46} \\
GR  & 1.47 & 1.85 & 1.83 & 1.32 & 1.59 & 1.36 & \textbf{1.28} & 1.44 \\
HR  & 1.93 & 2.21 & 2.14 & 1.77 & 1.83 & 1.89 & \textbf{1.72} & 1.74 \\
HU  & 1.66 & 1.86 & 1.75 & 1.59 & 1.61 & 1.40 & \textbf{1.36} & 1.45 \\
IE  & 1.78 & 1.57 & 1.53 & 1.48 & 1.46 & \textbf{1.42} & 1.44 & 1.46 \\
IS  & 1.19 & 1.21 & 1.17 & 1.19 & 1.11 & 1.10 & \textbf{1.09} & 1.14 \\
IT  & 1.62 & 1.69 & 1.51 & 1.63 & \textbf{1.09} & 1.28 & 1.18 & 1.47 \\
LT  & 1.40 & 1.91 & 1.80 & 1.34 & 1.41 & 1.37 & \textbf{1.32} & 1.64 \\
LU  & 1.88 & 2.28 & 1.92 & 1.94 & 1.56 & 1.56 & 1.51 & \textbf{1.27} \\
LV  & 1.68 & 1.56 & 1.50 & 1.60 & 1.51 & \textbf{1.40} & 1.49 & 1.64 \\
ME  & 2.22 & 2.27 & 2.19 & 1.98 & \textbf{1.95} & \textbf{1.95} & 2.04 & 2.14 \\
MK  & 3.61 & 3.50 & 3.31 & 3.13 & 3.22 & 3.22 & 3.09 & \textbf{2.77} \\
NL  & 1.52 & 1.76 & 1.66 & 1.54 & \textbf{1.22} & 1.59 & 1.40 & 1.39 \\
NO  & 2.05 & 2.15 & 2.07 & 2.01 & 1.79 & 1.84 & 1.77 & \textbf{1.54} \\
PL  & 1.27 & 1.95 & 1.84 & 1.27 & \textbf{1.15} & 1.18 & 1.18 & 1.27 \\
PT  & 1.39 & 1.73 & 1.63 & 1.18 & \textbf{1.11} & 1.22 & 1.14 & 1.25 \\
RO  & 1.30 & 1.66 & 1.71 & 1.24 & 1.16 & 1.11 & 1.10 & \textbf{1.06} \\
RS  & 1.74 & 1.55 & 1.55 & 1.60 & 1.50 & 1.53 & 1.58 & \textbf{1.46} \\
SE  & 1.73 & 2.05 & 1.91 & 1.76 & 1.64 & 1.69 & 1.65 & \textbf{1.39} \\
SI  & 1.78 & 2.31 & 2.31 & 1.86 & 1.60 & 1.65 & 1.55 & \textbf{1.40} \\
SK  & 1.25 & 1.25 & 1.17 & 1.18 & \textbf{1.08} & 1.09 & 1.09 & 1.11 \\
TR  & 1.37 & 1.47 & 1.35 & 1.29 & 1.15 & 1.29 & \textbf{1.14} & 1.31 \\ 
\hline
\end{tabular}
\label{tabP}
\end{center}
\end{table}

\begin{figure}[h]
	\centering
	\includegraphics[width=0.24\textwidth]{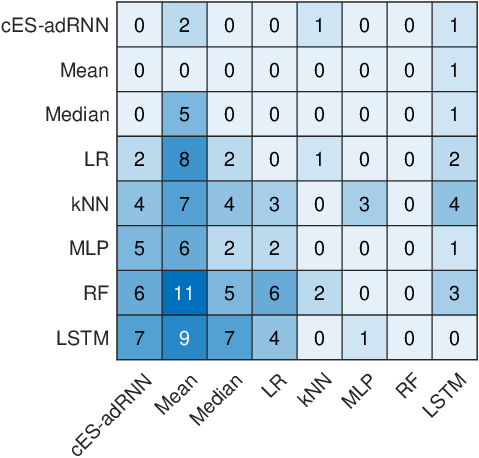}
        \caption{Results of the Diebold-Mariano tests.} 
	\label{fig4}
\end{figure}

A Diebold-Mariano test \cite{Die95} was conducted to assess the statistical significance of the differences between the forecasts generated by each pair of models based on individual country errors.
The results are presented in Fig. \ref{fig4}. The diagram illustrates the number of times the model shown on the y-axis is statistically more accurate than the model shown on the x-axis. 
For instance, in the last row of the diagram, we observe that LSTM outperformed cES-adRNN for 7 countries, outperformed Mean for 9 countries, outperformed Median for 7 countries, etc. From the figure, it is evident that the best-performing models were RF, LSTM, and kNN. These models were found to be more accurate than other models 33, 28, and 25 times, respectively. At the same time, they were outperformed by 0, 13, and 4 other models, respectively. 
On the other hand, the Mean method exhibited the poorest performance, winning only once and being outperformed 48 times. It is worth noting that RF was the only model that was not outperformed by any other model in the evaluation.

Fig. \ref{fig2} illustrates examples of forecasts for selected countries and test points.
It is important to note that even in situations where the base forecasts exhibit a substantial dispersion, the meta-models are capable of generating forecasts that closely align with the target values.
This is particularly evident in scenarios where the majority of base models produce forecasts that deviate significantly from the target value, such as test point 25 for PL, ES, and GR data. In contrast, the Mean and Median methods tend to follow the majority and produce inaccurate predictions in such cases.

It is worth noting that LSTM was able to achieve forecasts close to the target value despite the fact that no base model even came close to it, see test point 94 for FR and 99 for GB.
It was probably helped by the information from the recent past contained in the internal states of LSTM cell. 
Other meta-models do not use such information. To test the ability of the meta-models to produce forecasts outside the interval of the base models' forecasts (let us denote this interval for the $i$-th test point by $Z_i$), we counted the number of such cases out of the 3500 forecasts produced by each meta-model. The results are shown in row $N_1$ of Table \ref{tabN}. Row $N_2$ shows how many of these $N_1$ cases concern the situation where the target value also lay outside the Z-interval, on the same side as the meta-model forecast. Row $N_3$ shows the number of cases out of $N_1$ for which the meta-model generates more accurate predictions than the Median method. 

It is evident from Table \ref{tabN} that LSTM generates far more forecasts outside of $Z_i$ than other models. This may indicate better extrapolation properties of LSTM. The second model with the highest $N1$ was MLP, while RF least frequently went outside of $Z_i$.
However, when comparing LSTM and RF specifically on the 447 cases where LSTM fell outside of $Z_i$, it is worth noting that LSTM was more accurate than RF in less than half of these cases. This finding suggests that LSTM's high flexibility and extrapolation capacity do not necessarily translate into improved accuracy. Moreover, it can increase the risk of overfitting.

\begin{table}[htbp]
\setlength{\tabcolsep}{7pt}
\caption{Extrapolation properties of the meta-models.}
\begin{center}
\begin{tabular}{|c|c|c|c|c|c|c|c|}
\hline
 & Mean & Median & LR & kNN  & MLP  & RF & LSTM \\ 
\hline
$N_1$ & 0 & 0 & 48 & 108 & 150 & 34 & 447 \\
$N_2$ & 0 & 0 & 13 & 60  & 58  & 18 & 192 \\
$N_3$ & 0 & 0 & 27 & 73  & 75  & 23 & 244
\\ 
\hline
\end{tabular}
\label{tabN}
\end{center}
\end{table}




\section{Conclusions}

Combining forecasts has been widely recognized as a method to enhance forecast accuracy and robustness by integrating the available information from individual forecasts. This has been demonstrated in numerous papers and forecasting competitions. While averaging is the most commonly used method of combining, requiring no additional training and being computationally efficient, our study reveals that stacking, which involves meta-learning on forecasts generated by multiple models, can offer even greater benefits.

In our experimental study, which focused on forecasting time series with multiple seasonal patterns, we observed that meta-models of various types consistently outperformed the Mean and Median methods of combining across the majority of cases. Notably, the non-linear models such as kNN, MLP, RF, and LSTM exhibited higher accuracy compared to the linear regression model. Among the meta-models, RF stood out by generating the most accurate predictions and displaying little sensitivity to the size of the training set. Our findings highlight the superiority of meta-models in capturing complex patterns in time series and their ability to enhance forecasting performance in challenging scenarios.

Future research will focus on the development of advanced ML models specifically tailored for time series and sequential data to enhance the predictive capabilities of combining forecasts in bagging and boosting scenarios. 

\section*{Acknowledgment}

The author wishes to thank Slawek Smyl and Paweł Pełka for providing forecasts from the base models.


\end{document}